%% file: iccp26_main.tex
\documentclass[10pt,journal,compsoc]{IEEEtran}
\newif\ifpeerreview

\peerreviewfalse

\input{preamble}

\usepackage[nocompress]{cite}
\usepackage{url}
\usepackage{amsmath,amssymb,graphicx}

\usepackage[switch]{lineno}

\newcommand{\paperID}{XXXX}

\title{LFD: Enabling Real-World Lensless Face Recognition with a Large-Scale Dataset}

\author{Junho Kim*, Salman S. Khan*, Sara Wan, Tomi Kuye, Ashok Veeraraghavan
\IEEEcompsocitemizethanks{\IEEEcompsocthanksitem J. Kim, S.S. Khan, S. Wan, T. Kuye, A. Veeraraghavan are with the Department of Electrical and Computer Engineering, Rice University, Houston, TX, 77005.\protect\\
E-mail: jk84@rice.edu, sk135@rice.edu, av21@rice.edu \protect\\
*J. Kim and S. S. Khan contributed equally to this work}

}

\begin{document}

\IEEEtitleabstractindextext{%
\begin{abstract}
Face recognition is a ubiquitously used computer vision task that has a wide range of applications ranging from everyday smartphone biometrics to high-stakes security systems. Most face recognition systems rely on traditional cameras, which often suffer from limitations such as bulky form factors, high costs, and limited privacy protection. To address these limitations, lensless cameras have emerged as an alternative. Lensless cameras use thin optical encoders instead of lenses, enabling smaller size, lower cost, and greater design flexibility. These cameras are typically paired with reconstruction algorithms that convert raw captures into recognizable images. However, reconstructed images often contain artifacts, and the reconstruction methods struggle to generalize well to real-world conditions. Furthermore, existing face datasets do not account for the artifacts present in lensless images. To address this issue, we introduce the \datasetfull (\datasetshort), a large-scale, real-world lensless face dataset. \datasetshort comprises 21,080 lensless raw measurements, reconstructions, and standard images of faces captured under diverse lighting, angle, and distance. Our key contributions are: (1) Real-world lensless face data: \datasetshort focuses on capturing a diverse face dataset with varying levels of artifacts introduced under different environments; (2) In-the-wild captures: 4,976 images are captured in outdoor settings with varying intensities of natural light and different background patterns; (3) Multiple lensless devices: \datasetshort includes face images collected from three different types of lensless cameras, each with a unique optical encoder mask. We use this hardware diversity to demonstrate generalization across different lensless cameras. Through comprehensive evaluations and data attribute analysis, we show that \datasetshort effectively captures shared features and artifacts across different lensless imaging devices, making it a valuable dataset for advancing lensless face recognition. Our dataset and code are available at \texttt{https://jk-junhokim.github.io/lfd\_lensless\_face\_recog/}.
\end{abstract}

}

\maketitle

\IEEEraisesectionheading{
  \section{Introduction}\label{sec:introduction}
}
%
%
%
%
\IEEEPARstart{F}{ace} recognition has become deeply integrated into daily life with applications such as unlocking smart devices. Beyond personal devices, face recognition is increasingly applied in areas such as robotics, security, and surveillance~\cite{zhao2003face}. Driven by this expanding range of applications, extensive research has focused on leveraging trainable machine learning algorithms using facial data. As a result, a wide range of specialized face recognition models has been developed to address these diverse applications and scenarios~\cite{deng2017marginal, deng2019arcface, yang2020fan}. Most embedded face recognition systems rely on conventional camera modules. However, as applications increasingly demand smaller, lighter, and more cost-effective solutions, traditional lens-based cameras struggle to meet these requirements. This growing tension between miniaturization and imaging performance has become a central imaging challenge.


\begin{figure}[ht]
  \centering
  \includegraphics[width=0.85\columnwidth]{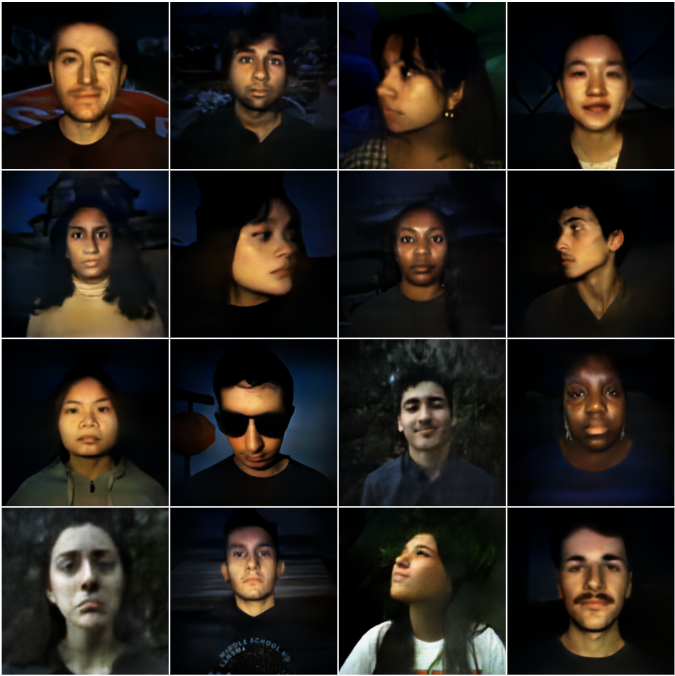}
  \caption{\textbf{Examples from Lensless Face Dataset.} Featuring subjects across diverse skin tones, expressions, poses, illumination conditions, occlusions, distances, and scene backgrounds.}
  \label{fig:lfd_examples}
\end{figure}

Lensless cameras have emerged as a lightweight, low-cost alternative by replacing bulky lenses with ultra-thin optical encoders~\cite{asif2016flatcam, boominathan2020phlatcam, antipa2017diffusercam}. Without lenses, these cameras capture raw multiplexed measurements with no local structures intact.
Efforts to reconstruct lensless images have led to a range of methods, from convolutional neural networks (CNNs)~\cite{khan2020flatnet}, physics-based neural networks~\cite{monakhova2019learned} to diffusion models~\cite{verma2025diffusion, cai2024phocolens}, most of which are trained and evaluated on simulated or display-captured data. Despite advances in reconstruction algorithms, current methods still struggle with artifacts like blurriness, color distortion, and vignetting which are unique to lensless imaging modalities. These artifacts have direct consequences on face recognition/verification performance~\cite{tan2018face}, and therefore need to be accounted for prior to the development of these inference algorithms. 
This underscores the urgent need for real-world lensless datasets for face recognition, enabling the training of deep learning models that support the practical deployment of lensless cameras in applications with compact form factor and weight constraints.

To address the lack of real-world data, we introduce the \datasetfull(\datasetshort), a collection of lensless face images that capture real-world conditions essential for training and evaluating lensless face recognition models. We will release the dataset upon acceptance of the paper. The main contributions of our work are as follows:
\begin{itemize}[leftmargin=1.2em]
  \item \textbf{Comprehensive Face Dataset:} We introduce \datasetshort, a large-scale, diverse dataset comprising 21,080 lensless raw measurements, reconstructions, and standard webcam face images. Our dataset is designed to capture real-world variations in face poses, expressions, lighting conditions, and scale.

  \item \textbf{Wide Range of Real-World Conditions:} \datasetshort is the first lensless face dataset to include both \textit{controlled} indoor environments and \textit{in-the-wild} outdoor settings. This spatial diversity introduces variations in both artificial indoor lighting and natural outdoor lighting. Also, lensless artifacts are naturally introduced from the diverse conditions including small reflective accessories, different backgrounds, low-lighting, providing a realistic environment for camera deployment. Additionally, our dataset includes subject skin tone annotations.

  \item \textbf{Generalization Across Mask Designs and Prototypes:} \datasetshort primarily comprises 19,100 images captured using PhlatCam (Prototype 1). It also includes 1,272 images from a second PhlatCam (Prototype 2) and 708 images from a Random Binary lensless camera. We show that face recognition models trained on Prototype 1 data generalize well across both similar mask types (Prototype 2) and fundamentally different mask designs (Random Binary). We are the first to present a data-driven comparison of both types of generalization.
\end{itemize}

\section{Related Work}
\subsection{Mask-based Lensless Imaging} Our study focuses on imaging systems using mask-based lensless cameras, where traditional focusing lenses are replaced with ultra-thin optical encoders placed closely (approximately 1.95\,mm) in front of CMOS sensors. This design significantly reduces the camera form factor, weight, and cost. Several lensless camera prototypes have been developed over the years, each employing different mask designs. For example, FlatCam~\cite{asif2016flatcam} uses a binary coded amplitude mask, DiffuserCam~\cite{antipa2017diffusercam} uses a caustic pattern-based phase mask, PhlatCam~\cite{boominathan2020phlatcam} uses a contour pattern-based phase mask, and a Random Binary camera uses a dotted random binary mask. Different mask designs aim to balance light efficiency and design flexibility, ensuring sufficient scene information is captured while allowing control over the mask-to-sensor distance for compact and flexible system design. 

Prior work has demonstrated the potential of mask-based lensless cameras in various applications. Mask-based lensless systems have been used for 3D imaging~\cite{antipa2017diffusercam}, microscopy~\cite{adams2022vivo}, optical encryption~\cite{khan2024opencam}, and gaze estimation~\cite{jain2025flattrack}. 

\input{assets/tables/dataset_comparison}

\subsection{Face Recognition} The field of face recognition has matured significantly in recent years, driven by its expanding role in a wide range of applications. At its core, face recognition involves determining whether two facial images belong to the same individual. Deep learning models are trained to represent faces as high-dimensional feature vectors, or embeddings, that capture distinctive facial attributes. Modern methods leverage CNNs to extract discriminative facial features~\cite{parkhi2015deep}, often enhanced with techniques like 3D face modeling~\cite{taigman2014deepface} and additive angular margin loss~\cite{deng2019arcface}. However, such methods require standard images and are ineffective when directly applied to lensless face images. Thus, we propose a data-driven approach to train face recognition models for improved performance on lensless face images. Features extracted from face recognition models can then be used to perform face verification. Widely used face recognition and verification datasets like VGGFace2~\cite{cao2018vggface2}, Labeled Faces in the Wild~\cite{huang2008labeled}, and MS-Celeb1M~\cite{guo2016ms} do not capture the unique characteristics and artifacts introduced by lensless imaging systems. Therefore, there is a need for a large-scale, diverse lensless face dataset along with thorough experimental and attribute-level analyses to assess real-world applicability.

\subsection{Lensless Face Recognition} 
Tan et al.~\cite{tan2018face} investigated lensless face recognition using FlatCam~\cite{asif2016flatcam}, and compared its performance against that of standard images. While this work demonstrated the early feasibility of lensless face recognition, it relied on simulated training data. The evaluation dataset they collected, the FlatCam Face Dataset (FCFD), was captured under controlled conditions, including close-range imaging, strong lighting, and minimal pose variation. These settings preserve more light and spatial information at the sensor, aiding reconstruction but failing to reflect real-world challenges. In contrast, we introduce the \datasetshort, which features higher-quality reconstructions while preserving diverse and challenging real-world conditions as shown in Fig.~\ref{fig:lfd_examples}. In recent years, lensless cameras have attracted attention due to their privacy-preserving characteristics, with methods proposed for face recognition~\cite{henry2023privacy, cai2024lenslessface}. In this work, however, we focus on developing a robust, large-scale dataset for lensless face recognition to enable novel algorithm development and real-world deployment.


\section{LFD: Lensless Face Dataset}

\subsection{Prototype Overview} The main contribution of our work is the \datasetfull (\datasetshort). Among various lensless mask designs, PhlatCam has demonstrated strong reconstruction quality by effectively balancing light efficiency and design flexibility~\cite{boominathan2020phlatcam}. For this reason, we use PhlatCam-based lensless cameras. We will refer to this prototype as Prototype 1. Additionally, to verify the effectiveness of \datasetshort for transfer learning, we collect test sets using (a) Prototype 2, which differs from Prototype 1 in terms of mask fabrication errors -- it has the same mask design but the difference in performance is due to fabrication and assembling error and (b) random binary camera that uses a random dotted point spread function (PSF) pattern similar to that of lensless cameras proposed in~\cite{reshetouski2020lensless,fu2022diffractive}. Prototype 1 serves as the primary device for data collection from 104 participants (19,100 images), while Prototype 2 is used to collect test data from an additional 7 subjects (1,272 images). Random binary camera was used to collect test data from 4 subjects, yielding 708 images (details in Table~\ref{tab:facedatasets}). \textit{All data collection procedures were approved by the Institutional Review Board, and informed consent was obtained from all participants prior to data collection.}

\begin{figure}[t]
  \centering
  \includegraphics[width=0.90\columnwidth]{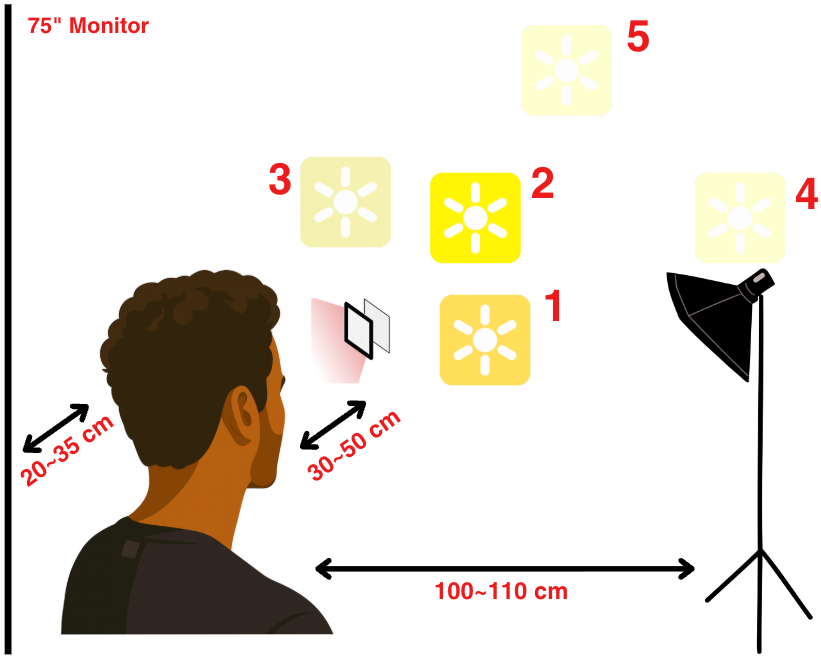}
  \caption{\textbf{Indoor Data Collection Setup.} We use five light sources: (1) right, (2) front, (3) left, (4) far right, and (5) ceiling. For outdoors, the setup remains the same, but no artificial lighting and monitor screen are used.}
  \label{fig:camerasetup}
\end{figure}


\subsection{Indoor ``Controlled'' Captures} Indoor data collection was conducted in a fixed laboratory setting, where we artificially introduced variations in illumination and background. To create different illumination conditions, we used five light sources with varying intensity, color temperature, angle, and distance from the subject (see Fig.~\ref{fig:camerasetup}). Subjects were seated between the camera and a television, which displayed random background images from the Microsoft COCO dataset~\cite{lin2014microsoft}, excluding those with human faces.

\begin{figure*}[t]
  \centering
  \includegraphics[width=0.95\linewidth]{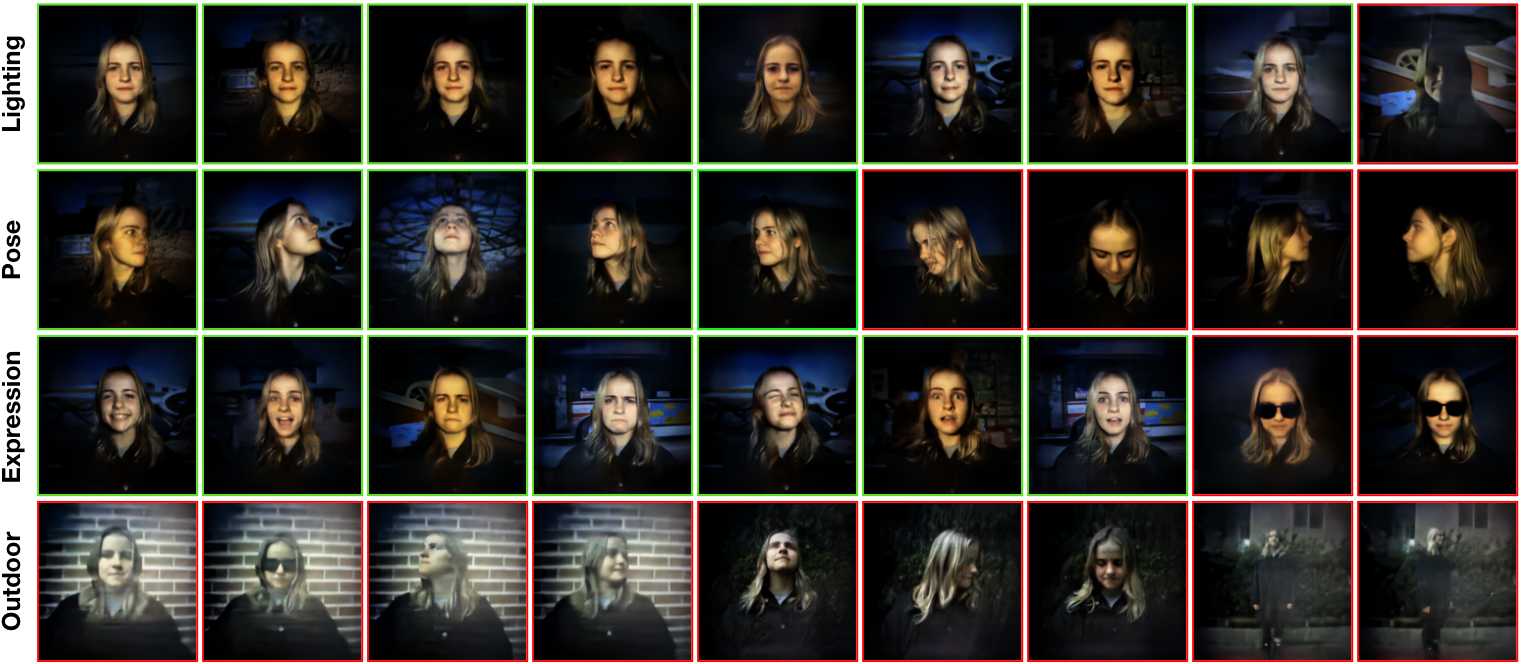}
  \caption{\textbf{Lensless Face Dataset.} Reconstructed images of a single subject captured using Prototype 1, organized by lighting, pose, expression, and outdoor conditions (bottom row). We define five face verification subsets: \textit{Indoor Easy} (green border) with good lighting, simple expressions, and frontal poses; \textit{Hard} (red border) with low-light or outdoor lighting, occlusion, extreme poses, and longer subject distances; \textit{Complete} (green + red border) with all available conditions; \textit{Indoor All} with all indoor conditions; and \textit{Outdoor} covering uncontrolled natural outdoor scenes. Outdoor samples are shown in the bottom row.}
  \label{fig:capturepalette}
\end{figure*}

Each subject was recorded under 9 different lighting conditions, created by various combinations of five light sources. For each light combination, subjects rotated their heads across 10 evenly spaced angles (approximately 35° apart), including extreme side-facing positions. They were also instructed to display 5 facial expressions (neutral, smiling, sad, winking, surprised) and to wear sunglasses to introduce occlusion. When wearing sunglasses, subjects tilted their heads downward to minimize reflective glare, which can cause saturation across the face. Reconstructed images of example captures are shown in Fig.~\ref{fig:lfd_examples} and Fig.~\ref{fig:capturepalette}. Details of the lighting combinations, experimental setup, and data collection software are provided in the supplementary material.

\begin{figure}[ht]
  \centering
  \includegraphics[width=0.85\columnwidth]{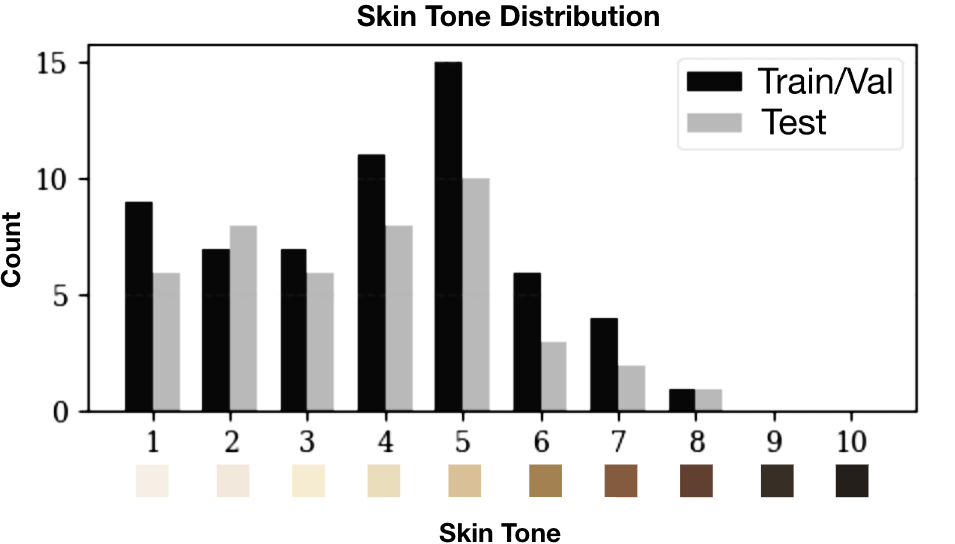}
  \caption{\textbf{Monk Skin Tone Distribution of LFD Participants.} Distribution of skin tones across the training/validation and test sets using the Monk Skin Tone scale. The randomized split maintains a balanced distribution across both sets, mitigating skin tone bias in evaluation.}
  \label{fig:skintone}
\end{figure}

\subsection{Outdoor ``In-the-Wild'' Captures} Outdoor captures were conducted following the indoor session. In some cases, participants returned on a different day, which introduced temporal variability. We selected two outdoor locations: one near a brick building wall and the other in front of dense foliage. Outdoor captures were also affected by time and weather, as the direction and intensity of sunlight influence raw measurements. We avoided overly cloudy or rainy days and collected data in the morning and afternoon to ensure sufficient sunlight.

For each location, we followed the same protocol as the indoor session (10 head poses, 5 expressions, and 1 occlusion). Additionally, we collected standing poses, where subjects stood 100–120 cm away from the camera and faced 6 different directions. This was done to test face recognition at extended distances (see Fig.~\ref{fig:capturepalette}, bottom row). Due to the multiplexing nature of lensless cameras, outdoor captures appear more influenced by the background and stray illumination. These real-world outdoor samples capture the challenges of using lensless cameras in natural environments, offering a more realistic and comprehensive assessment of face recognition.

\input{assets/tables/head_pose_distribution}

\subsection{Data Distribution}

\subsubsection{Skin Tone}
All data collection was supervised by the authors to ensure correct pose, illumination, and expression conditions. For each subject, we recorded metadata including height, weight, and skin tone, annotated using Monk's scale~\cite{monk2023monk}. Camera height and distance were adjusted based on subject height to ensure the face was roughly in the center of the field of view. Skin tone annotations are used to ensure a balanced distribution across training and test splits (see Fig.~\ref{fig:skintone}).

\subsubsection{Head Pose}
We further validate our dataset by analyzing variations across different head poses. Capturing clear pose differences is important for modeling unconstrained real-world scenarios and enabling generalization~\cite{zheng2018cross}. We estimated the head poses using the publicly available implementation~\cite{nafie_headpose_github} based on a Support Vector Regression model trained on facial landmarks extracted from the AFLW face dataset~\cite{koestinger2011annotated} using MediaPipe\cite{lugaresi2019mediapipe}. The AFLW dataset also contains associated head-pose ground truths. Poses are reported using Euler angles $(\theta_p, \theta_y, \theta_r)$ corresponding to pitch, yaw, and roll (in degrees), respectively. For each pose category $c$, we use the webcam images to compute the mean $\mu_c$ and standard deviation $\sigma_c$ over samples. As shown in Table~\ref{tab:pose_stats}, frontal poses exhibit near-neutral alignment. Upward and downward instances show positive and negative pitch, respectively, with minimal variation in yaw and roll. Lateral poses exhibit strong separation along the yaw axis, with left-facing instances showing positive yaw and right-facing instances showing negative yaw. These results indicate low intra-class variance and clear inter-class separation across head poses, with minimal overlap between different head pose categories. Comprehensive head pose distributions are visualized in the supplementary material.

\begin{figure*}[t]
  \centering
  \includegraphics[width=\linewidth]{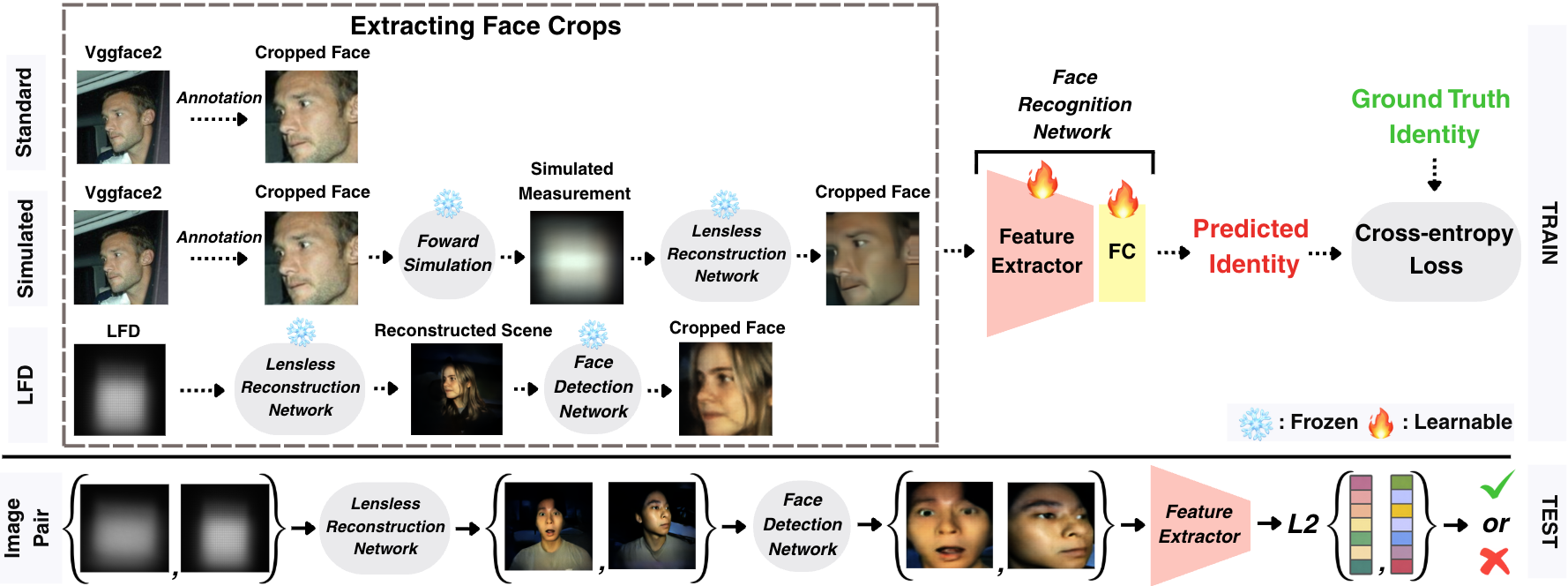}
  \caption{\textbf{Lensless Face Recognition Pipeline.} During training, face images are extracted and cropped separately from standard, simulated, and \datasetshort reconstructions, and are used to train a face recognition network. During testing, raw lensless measurements are first reconstructed, then passed through a face detection network to crop the face region, and finally through the frozen feature extractor to verify identity.}
  \label{fig:pipeline}
\end{figure*}

\section{Lensless Face Recognition Pipeline}
\label{sec:model}

\subsection{Lensless Forward Model} The lensless cameras used in our work share a common structure: an optical encoder mask placed directly in front of the image sensor, and an aperture positioned above the mask to constrain incoming light and ensure the resulting pattern fits within the sensor's dimensions. The uniqueness of each camera lies in its mask design. In the phase mask, light from a point source undergoes phase modulation as it passes through the mask, while in an amplitude mask, light is selectively passed or blocked before reaching the sensor. As a result, each point in the scene is projected as a unique pattern on the sensor, corresponding to the PSF of the lensless system. This property enables the lensless imaging process to be modeled using the following forward model:

\begin{equation}
  Y = P * X + N,
  \label{eq:fwdmodel}
\end{equation}

where \(Y\) is the sensor measurement, \(P\) is the PSF, \(*\) is the full-size convolution operator, \(X\) is the scene, and \(N\) is the additive noise. 

\subsection{Image Reconstruction} While the PSF uniquely encodes incoming light, the absence of a lens causes the raw measurements to become multiplexed projections of individual point sources. Image reconstruction involves solving an inverse problem to recover the original scene \(X\) from the multiplexed measurement \(Y\) and the PSF \(P\). While several reconstruction methods have been proposed, we use FlatNet~\cite{khan2020flatnet} due to its strong performance shown in lensless imaging applications. Compared to diffusion-based model~\cite{cai2024phocolens} which are computationally intensive and data hungry, FlatNet is a lightweight, two-stage model suitable for fast, real-time reconstruction. The first stage is the \textit{Trainable Camera Inversion}, which maps raw measurements \(Y\) to the standard image space. This operation is performed in the Fourier domain and is represented as $X_{\text{intermediate}} = \mathcal{F}^{-1} \left( \mathcal{F}(W) \odot \mathcal{F}(Y) \right)$. Here,
$X_{\text{intermediate}}$ is an intermediate estimate of the scene. The filter $W$ is initialized using the Wiener matrix ($W = \mathcal{F}^{-1} ( \frac{H^*}{|H|^2 + \Gamma})$) and is updated during training. $H$ is the PSF in the frequency domain, $\Gamma$ is the regularization parameter. The second stage is the \textit{Perceptual Enhancement Network}, which uses U-Net~\cite{ronneberger2015u} as the backbone to enhance the visual quality of \(X_{\text{intermediate}}\). Both stages are trained jointly in an end-to-end manner.

\subsection{Face Detection} Following the standard face verification pipeline, we apply Faster R-CNN~\cite{ren2015faster} face detection model to the FlatNet reconstructions. If a face is detected, we crop the image using the predicted bounding box to isolate the face and reduce background clutter, which improves verification accuracy. Otherwise, the full image is used.

\subsection{Face Recognition} 
Finally, we use SENet~\cite{hu2018squeeze} as the face recognition network. We train it using cross-entropy loss to classify identities, while all other pipeline stages remain frozen. At test time, the classification head is discarded and the network serves as a feature extractor, whose embeddings are compared using a similarity metric for identity verification (see Fig.~\ref{fig:pipeline}).


\section{Experiments and Results}
In this section, we present implementation details of our face recognition pipeline and a comprehensive evaluation of the \datasetshort as an effective training and evaluation dataset for current and future lensless face recognition systems. We chose simple, light weight, and widely used face detection and verification algorithms to ensure our results reflect differences in data rather than the choice of interchangeable methods.

\subsection{Implementation Details}
\textbf{Reconstruction Network:}
We train FlatNet separately for each lensless camera using measurements simulated from natural images in the MIRFLICKR dataset~\cite{huiskes2008mir}, following the Lensless Forward Model described in Section~\ref{sec:model}. Once trained, we apply each FlatNet to reconstruct both simulated and raw \datasetshort measurements. Since the dataset comprises three distinct lensless cameras, we train three separate FlatNet models accordingly.\\
\textbf{Face Detection Network:}
We train four versions of Faster R-CNN. For the webcam, we train a single network directly on standard WIDER FACE images~\cite{yang2016wider}. For each lensless camera, we first simulate lensless measurements from the WIDER FACE dataset using the corresponding PSF, then apply the pre-trained FlatNet to reconstruct these measurements. We use the resulting reconstructions to train the detection network for each camera. This yields three camera-specific FlatNet–Faster R-CNN pairs, each comprising a dedicated reconstruction and detection network.\\
\textbf{Face Recognition Network:}
We use SENet as implemented in~\cite{cao2018vggface2} using the official VGGFace2 training dataset. We train three separate SENet models: one on standard VGGFace2, one on reconstructed simulated measurements obtained from VGGFace2, and one on the reconstructed proposed \datasetshort training dataset. Prior to training the recognition network, we crop the face out. For standard and simulated data, we use the provided annotations to do this, while for the \datasetshort, we use Faster-RCNN trained for Prototype 1 (see Fig.~\ref{fig:pipeline} \textit{Extracting Face Crops}). The \datasetshort training dataset refers to the data captured using Prototype 1 corresponding to 60 random participants. Please note that the reconstruction, face detection and face recognition models are trained separately and \textbf{not} in an end-to-end fashion.

Face recognition models trained on standard, simulated, and the \datasetshort training data are hereafter referred to by their respective training data names. All models are trained for 60 epochs using the AdamW optimizer with an initial learning rate of $3 \times 10^{-4}$. These hyperparameters are consistent across all training settings. All models were trained and evaluated using NVIDIA GeForce RTX 2080 Ti GPU. Please note, we train the reconstruction, face detection, and face recognition models separately.

\input{assets/tables/prototype1_auc}
\input{assets/tables_supp/attribute_eval}

\subsection{Comparison of \datasetshort with Standard and Simulated Data}
In this section, we evaluate lensless face recognition models to highlight that the \datasetshort is the most effective training data for real-world lensless face verification. While standard and simulated approaches are easy to obtain, they fail to transfer well to real-world lensless images. We evaluate on the Prototype 1 test set, which consists of data from 44 subjects, using five verification subsets: Indoor Easy, Indoor All, Outdoor, Complete, and Hard, defined by lighting and head pose difficulty (see supplementary material for detailed subset groupings). For each subset, we generate 5,000 positive (same identity) and 5,000 negative (different identity) pairs.

We use the L2 distance to measure similarity between image embeddings and compute the Area Under the Receiver Operating Characteristic curve (AUC)~\cite{fawcett2006introduction} by varying the similarity threshold and plotting the resulting True Positive Rate (TPR) against the False Positive Rate (FPR). Table~\ref{tab:proto1_auc} shows that the LFD model outperforms those trained on standard and simulated data across all verification subsets. ROC curves are visualized in the supplementary material. In general, the performance of all the models on outdoor samples is worse than the performance on indoor samples. However, using the \datasetshort training set, the gap between performance on indoor and outdoor data is significantly reduced, highlighting the importance of our proposed dataset. The oracle numbers for standard face recognition, along with the standard deviation for all values in this table, are provided in the supplementary material.

\subsection{Lensless Face Recognition Across Attributes and Capture Conditions}
A key contribution of the \datasetshort is having both artificial and natural variability to represent real-world conditions. \datasetshort's attributes and capture conditions can be grouped into four categories (i.e., expression, occlusion, pose, and illumination) which influence face recognition performance. We evaluate the face verification model and report the area under the ROC curve (AUC) for different categories in Table~\ref{tab:attribute_eval}. Expression-related attributes exhibit relatively small performance drop in AUC values between the \textit{Sad} and \textit{Wink}. Occlusion offered by sunglasses doesn't lead to significant drop in the AUC compared to the expression attributes.  Although expressions vary, they are captured with subjects facing the camera, resulting in consistent frontal views and relatively smaller variation in AUC. In contrast, illumination introduces a larger gap, with \textit{Ceiling + Front + Right} achieving noticeably higher performance than \textit{Outdoor} and \textit{Far-right} conditions. As expected, illumination conditions that adequately illuminate the face from all directions significantly improve reconstruction quality whereas insufficient lighting leads to information loss. Consistent with prior work~\cite{zheng2018cross}, head pose also presents a substantial challenge for face recognition. In particular, extreme poses such as \textit{Profile Left} and \textit{Profile Right}, which expose only one side of the face and reduce visible facial information, result in noticeably lower AUC compared to the frontal (straight) pose.

\vspace{-1.mm}
\subsection{Comparison with Existing Dataset}
In this section, we compare the \datasetshort and FCFD~\cite{tan2018face}. We begin by outlining the qualitative advantages of the \datasetshort and validating these advantages through two quantitative experiments.

\subsubsection{Qualitative Comparison}
\begin{figure}[t]
  \centering
  \includegraphics[width=0.95\columnwidth]{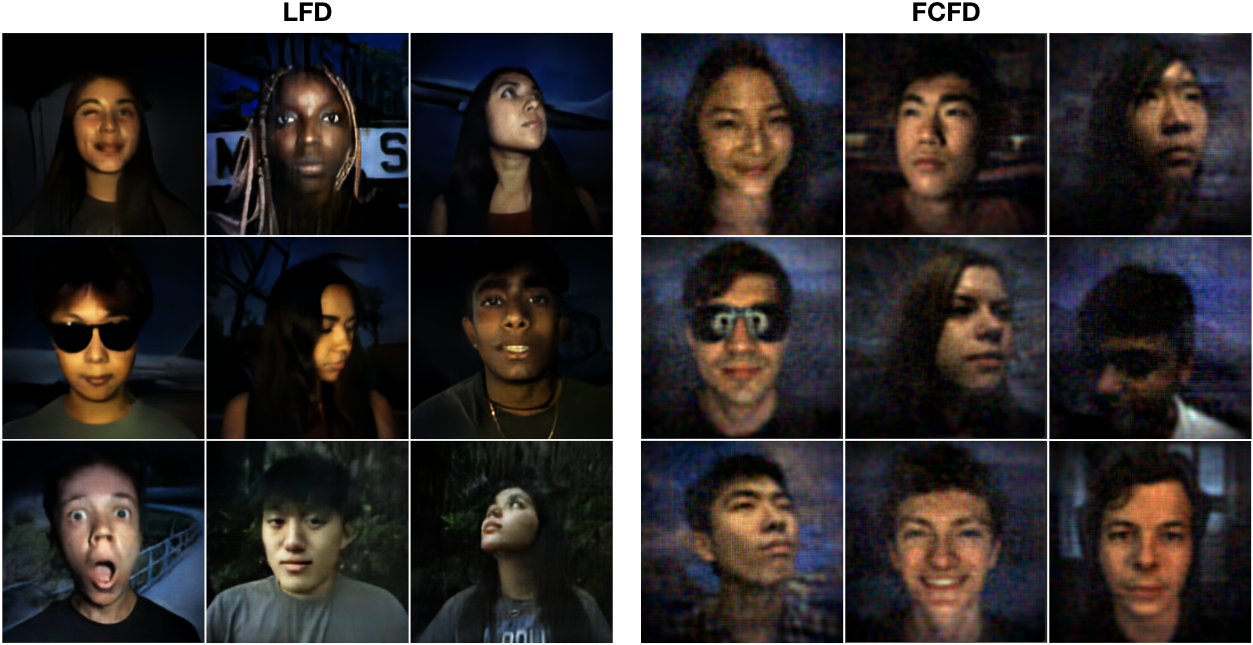}
  \caption{\textbf{Qualitative Comparison of \datasetshort and FCFD.} Compared to FCFD, \datasetshort reconstructions exhibit higher quality and cover more diverse and challenging capture conditions, including full profile poses, stronger expressions, varied illumination, outdoor scenes (bottom row), and longer subject distances. Zoom in for better visualization.}
  \label{fig:qualitative_comparison}
\end{figure}

To visually compare, we present examples in Fig.~\ref{fig:qualitative_comparison} highlighting \datasetshort's high-resolution reconstructions across diverse capture conditions. Unlike FCFD, \datasetshort includes outdoor scenes essential for uncontrolled environment evaluation, more challenging low-light and long-distance captures, and more extreme pose variations spanning full profile views, whereas FCFD is limited to three-quarter poses.

\subsubsection{Quantitative Comparison}

\input{assets/tables/perceptual_similarity}

\begin{figure}[t]
  \centering
  \includegraphics[width=0.90\columnwidth]{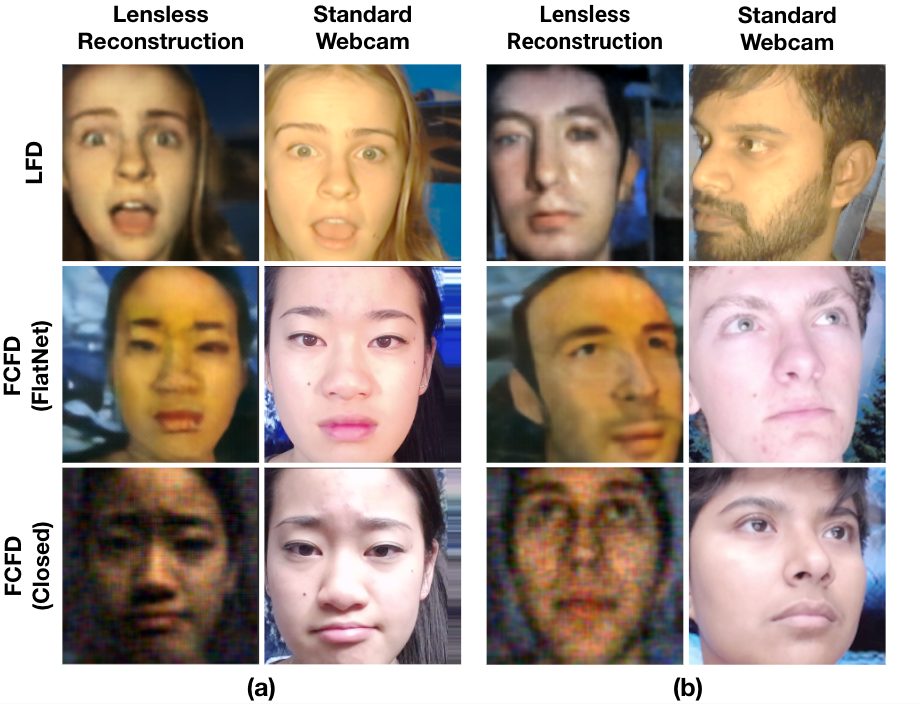}
  \caption{\textbf{Example Cross-modal Face Verification Pairs.} (a) Positive Pairs (images from same person) (b) Negative pairs (images from different people).}
  \label{fig:verificationpairs}
\end{figure}

In this experiment, we show that \datasetshort reconstructions are closer in quality to standard webcam images than those of FCFD. We measure this through two complementary metrics: (a) perceptual similarity metrics, and (b) a cross-modal face verification metric. We argue that since \datasetshort is captured using a phase-mask-based PhlatCam, its reconstructions exhibit a smaller domain gap with respect to standard lens-based images. This is desirable because models trained on standard face images can be more easily transferred or fine-tuned to lensless images, making lensless cameras more practical for real-world deployment. We describe each metric below.\\
\textbf{Perceptual Metrics:} First, we evaluate the perceptual similarity between lensless face reconstructions and their corresponding standard images captured with the webcam. We train FCFD models with 60 participants from FCFD, while following the LFD training pipeline in Fig.~\ref{fig:pipeline}. The reconstructions for FCFD include two types: a closed-form method from~\cite{tan2018face}, which is simple and fast but has severe artifacts, and FlatNet~\cite{khan2020flatnet}, a deep learning-based approach. Traditional image quality metrics such as PSNR and SSIM are unsuitable for our comparison due to spatial misalignments between lensless and standard captures. To address this, we apply face detection and use cropped faces, which significantly reduce misalignment. To evaluate similarity to standard images while addressing residual misalignment, we use perceptual similarity metrics including contextual similarity (CX)~\cite{mechrez2018contextual}, LPIPS~\cite{zhang2018unreasonable}, and DreamSim~\cite{fu2023dreamsim}. Higher CX and lower LPIPS and DreamSim scores indicate higher similarity between lensless and standard images. Evaluation is conducted on 5,000 \textit{Easy} lensless–standard pairs for both the \datasetshort and FCFD. This enables direct comparison despite differing dataset distributions. As shown in Table~\ref{tab:perceptualsim}, the \datasetshort pairs outperform both types of FCFD reconstructions across all three similarity metrics. \\
\textbf{Cross-modal Face Verification:} Perceptual similarity metrics compare high-level image embeddings extracted from pre-trained image recognition models. However, since they are not optimized for face recognition, they do not measure similarity in the feature space relevant to distinguishing identities. Thus, in our second quantitative evaluation, we compare \datasetshort against FCFD using a cross-modal face verification approach. In cross-modal face verification, each test pair consists of one lensless and one standard webcam image, rather than two images from the same modality. Some of these pairs are shown in Fig.~\ref{fig:verificationpairs}. The feature extractor is trained exclusively on standard face images and is used to embed both images in each pair. The resulting similarity score thus directly reflects how closely lensless reconstructions align with standard images in the face recognition feature space. The rest of the testing pipeline remains the same as described in Fig.~\ref{fig:pipeline}. Higher quality reconstructions that closely resemble standard images are expected to yield a higher TPR at the same FPR.

To compare model performance, we evaluate TPR at FPRs of 0.05\%, 0.1\%, 0.2\%, and 0.5\%. As shown in Table~\ref{tab:tpr_verification}, our model consistently outperforms FCFD-trained models, with the performance gap narrowing at higher FPRs. Since face recognition applications prioritize low FPRs, our model demonstrates a clear advantage in the range of interest for real-world applicability.

\input{assets/tables/image_quality_tpr_verification}

\input{assets/tables/prototype2_auc}
\input{assets/tables/randombinary_auc}

\subsection{Generalization Ability Across PhlatCam Devices}

In this section, we demonstrate that \datasetshort serves as effective training data for future PhlatCam prototypes by evaluating its generalization to Prototype 2. Prototype 2 has the same mask design as Prototype 1 but differs in two aspects: mask fabrication differences and assembly error appearing as mask rotations/translations. Despite the differences in PSFs, images reconstructed from Prototype 1 and Prototype 2 share similar lensless artifacts (see supplementary material). We evaluate the generalization ability of the LFD model (model trained on \datasetshort training set) by testing it on Prototype 2 data (7 subjects), using the same illumination-based subset variations as Prototype 1.

The AUC scores presented in Table~\ref{tab:proto2_auc} show that the LFD model consistently outperforms models trained on standard and simulated data across all verification subsets, following the same trend observed for Prototype 1.

\subsection{Generalization Ability Across Different Lensless PSFs}

In this section, we demonstrate the effectiveness of the \datasetshort as training data for unseen lensless cameras that have mask patterns different from those used in PhlatCams. While the \datasetshort has shown strong real-world performance, high reconstruction quality, and robust generalization across different PhlatCam variants, future lensless cameras may differ significantly. To assess the future applicability of the \datasetshort, we evaluate it on the Random Binary camera, which features a fundamentally different mask design. We build this prototype by developing a phase mask corresponding to random binary dots PSF similar to the ones used in \cite{reshetouski2020lensless, fu2022diffractive}. We evaluate the performance on the Indoor easy subset that contains 735 positive and 735 negative pairs.

As shown in Table~\ref{tab:randombinary_auc}, the model trained on the \datasetshort outperforms all different models, including standard, simulated Random Binary, and FCFD-trained models. The FCFD models, trained on data from 60 participants in the FCFD dataset, are used to evaluate how well they generalizes to unseen lensless images compared to LFD. The LFD model demonstrates strong generalization to a completely different mask design, achieving significant performance gains over other models. We believe the \datasetshort holds further potential for future lensless camera designs.

\section{Conclusion}
This work presents the \datasetfull (\datasetshort), a new dataset tailored for lensless face recognition. Unlike prior datasets, \datasetshort captures real-world conditions including natural outdoor lighting, diverse backgrounds, extreme illumination scenarios, and a wide range of pose variations, comprising 21,080 samples of raw lensless measurements, reconstructed images, and corresponding webcam images. Through attribute analysis, qualitative comparisons, and quantitative evaluations, we show that \datasetshort offers higher quality and greater diversity than existing lensless face datasets. Models trained on \datasetshort outperform those trained on standard and simulated data for real-world lensless face recognition, and generalize across PhlatCam prototypes as well as cameras with entirely different designs, such as the Random Binary camera. This makes \datasetshort a valuable resource for advancing robust face recognition algorithms for lensless cameras, independent of any particular mask design. Furthermore, we demonstrate the feasibility of deploying lensless cameras in unconstrained real-world environments. Future work can explore face recognition algorithms developed using \datasetshort that directly operate on raw lensless measurements, exploiting their unique structure to bypass the reconstruction step entirely.

\ifpeerreview \else
\section*{Acknowledgments}
The authors would like to thank Vivek Boominathan, Jasper Tan, and Bhargav Ghanekar for their valuable feedback throughout the data collection process.
\fi

\bibliographystyle{IEEEtran}
\bibliography{references}

\ifpeerreview \else


\begin{IEEEbiography}[{\includegraphics[width=1in,height=1.25in,clip,keepaspectratio]{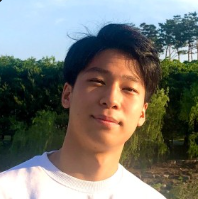}}]{Junho Kim} received the B.S. degree in Computer Science from Rice University in 2026. His research focus lies in computer vision, computational imaging, and generative models. He is currently a software engineer at ASML, San Jose, CA, USA.
\end{IEEEbiography}

\begin{IEEEbiography}[{\includegraphics[width=1in,height=1.25in,clip,keepaspectratio]{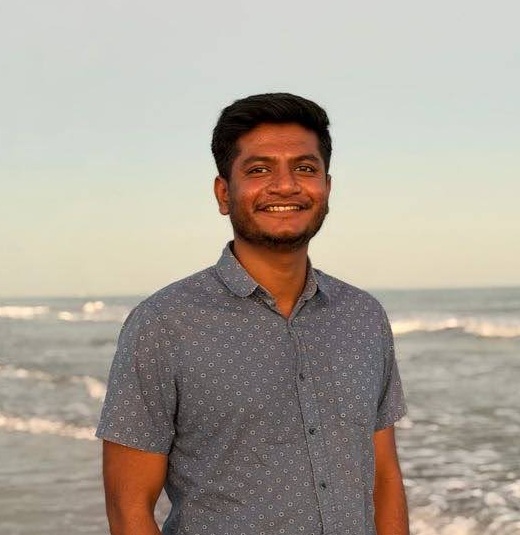}}]{Salman S. Khan}
received his B.S. degree in Electronics and Instrumentation Engineering from the National Institute of Technology, Rourkela in 2018, and his M.S. and Ph.D. degrees in Electrical Engineering from the Indian Institute of Technology Madras in 2023. He is currently a postdoctoral researcher in the Department of Electrical and Computer Engineering at Rice University, Houston, TX, USA. He is interested in computational imaging and digital health.
\end{IEEEbiography}

\begin{IEEEbiography}[{\includegraphics[width=1in,height=1.25in,clip,keepaspectratio]{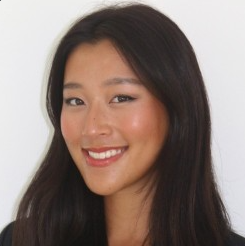}}]{Sara Wan} is currently pursuing her B.S. degree in Computer Science from Rice University, Houston, TX. Her research interests lie in machine learning and computational imaging.
\end{IEEEbiography}

\begin{IEEEbiography}[{\includegraphics[width=1in,height=1.25in,clip,keepaspectratio]{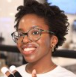}}]{Tomi Kuye} received the B.S. degree in Electrical and Computer Engineering from Rice University in 2026. Her research interests lie in lensless imaging and digital health.
\end{IEEEbiography}

\begin{IEEEbiography}[{\includegraphics[width=1in,height=1.25in,clip,keepaspectratio]{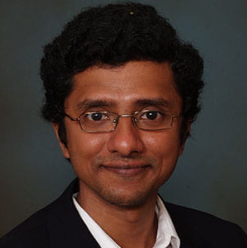}}]{Ashok Veeraraghavan} received the bachelor’s degree in electrical engineering from the Indian Institute of Technology, Madras, Chennai, India, in 2002 and the M.S. and Ph.D. degrees from the Department of Electrical and Computer Engineering, University of Maryland, College Park, MD, USA, in 2004 and 2008, respectively. He is currently Professor of Electrical and Computer Engineering, Rice University, Houston, TX, USA.
\end{IEEEbiography}





\fi

\end{document}

%% file: preamble.tex
\usepackage[table,dvipsnames]{xcolor}
\usepackage{tabularx}
\usepackage{multirow}
\usepackage{booktabs}
\usepackage{makecell}
\usepackage{amssymb}      
\usepackage{pifont}       
\usepackage{siunitx}
\usepackage{enumitem}
\usepackage{amsmath}

\usepackage{xspace}

\newcommand{\datasetfull}{\textit{Lensless Face Dataset}\xspace} 
\newcommand{\datasetshort}{\textit{LFD}\xspace} 

\newcommand{\best}[1]{\textbf{#1}}

%% file: assets/tables/dataset_comparison.tex
\begin{table*}
  \centering
  \caption{\textbf{Details of Lensless Face Datasets.} A comparison between the LFD proposed in this paper and the existing FCFD.}
  \label{tab:facedatasets}
  {\small
  \renewcommand{\arraystretch}{1.4}
  \setlength{\tabcolsep}{4pt}
  \begin{tabularx}{\textwidth}{@{}l c X c c c c c@{}}
    \toprule
    Dataset & 
    \makecell[c]{Image\\Type} & 
    \makecell[c]{Source Device} & 
    \makecell[c]{Image\\Resolution} & 
    \makecell[c]{Number\\of Images} & 
    \makecell[c]{Indoor/\\Outdoor} & 
    \makecell[c]{Number\\of Subjects} & 
    Distance \\

    \specialrule{0.2pt}{1pt}{1pt}

    \textbf{LFD (Ours)} & 
    \makecell[c]{Lensless\\+\\Standard} & 
    \makecell[c]{PhlatCam Prototype 1\\PhlatCam Prototype 2\\Random Binary Camera\\Logitech 270HD Webcam} & 
    \makecell[c]{640 × 640\\640 × 640\\520 × 520\\640 × 480} & 
    \makecell[c]{19,100\\1,272\\708\\21,080} & 
    Both & 
    \makecell[c]{104\\7\\4\\115} & 
    30--110 cm \\

    \specialrule{0.2pt}{1pt}{1pt}

    \makecell[l]{FCFD~\cite{tan2018face}} &
    \makecell[c]{Lensless\\+\\Standard} & 
    \makecell[c]{FlatCam\\Logitech C930e Webcam} & 
    \makecell[c]{256 × 256\\$\sim$1000 × 1000} & 
    24,112 & 
    Indoor Only & 
    88 & 
    15--38 cm \\

    \bottomrule
  \end{tabularx}
  }
\end{table*}

%% file: assets/tables/head_pose_distribution.tex
\begin{table}[t]
  \centering
  \caption{\textbf{Head Pose Statistics in \textit{LFD}.} Mean ($\mu$) and standard deviation ($\sigma$) of Euler angles (pitch, yaw, roll in degrees) for each pose category.}
  \label{tab:pose_stats}
  \renewcommand{\arraystretch}{1.35}
  \setlength{\tabcolsep}{8pt} 
  \begin{tabular*}{\columnwidth}{@{\extracolsep{\fill}}lccc@{}}
    \toprule
    \textbf{Pose} & $\boldsymbol{\theta_p}$ (Pitch) & $\boldsymbol{\theta_y}$ (Yaw) & $\boldsymbol{\theta_r}$ (Roll) \\
    \midrule
    Straight (Frontal) & 0.2 (4.4)   & 6.6 (8.2)   & 2.1 (4.0) \\
    Up                 & 24.2 (1.5)  & 5.4 (3.0)   & 1.1 (2.0) \\
    Down               & -25.7 (9.6) & -2.8 (2.1)  & 1.5 (1.3) \\
    Left               & -7.8 (7.9)  & 63.3 (2.6)  & -17.8 (9.4) \\
    Right              & -11.4 (9.7) & -58.4 (3.6) & 19.4 (7.9) \\
    \bottomrule
  \end{tabular*}
\end{table}

%% file: assets/tables/prototype1_auc.tex
\begin{table}[t]
  \centering
  \caption{\textbf{AUC Scores for Prototype 1 Evaluation.} The model trained on the LFD training set outperforms all others across all evaluation subsets.}
  \label{tab:proto1_auc}
  \renewcommand{\arraystretch}{1.35}
  \setlength{\tabcolsep}{5pt}
  \begin{tabular*}{\columnwidth}{@{\extracolsep{\fill}}lccccc@{}}
    \toprule
    \textbf{Model} & \makecell{\textbf{Indoor}\\\textbf{Easy}} & \makecell{\textbf{Indoor}\\\textbf{All}} & \textbf{Outdoor} & \textbf{Complete} & \textbf{Hard} \\
    \midrule
    Standard   & 0.884 & 0.830 & 0.625 & 0.763 & 0.599 \\
    Simulated  & 0.906 & 0.855 & 0.660 & 0.798 & 0.626 \\
    LFD (Ours) & \textbf{0.946} & \textbf{0.908} & \textbf{0.808} & \textbf{0.851} & \textbf{0.778} \\
    \bottomrule
  \end{tabular*}
\end{table}

%% file: assets/tables_supp/attribute_eval.tex
\begin{table}[t]
\centering
\caption{\textbf{LFD Evaluation Across Attributes and Capture Conditions.} Face verification AUCs highlight that illumination and extreme head poses introduce larger performance degradation compared to expression variations.}
\label{tab:attribute_eval}
\small
\renewcommand{\arraystretch}{1.2}
\setlength{\tabcolsep}{6pt}

\begin{tabular}{llc}
\toprule
Category & Attribute & AUC \\
\midrule
\multirow{5}{*}{\shortstack{Expression \& \\ Occlusion}}
& \best{Sad}            & \best{0.9641} \\
& Smile                 & 0.9624 \\
& Surprised             & 0.9557 \\
& Wink                  & 0.9533 \\
& Sunglasses            & 0.9370 \\
\midrule
\multirow{11}{*}{Pose}
& \best{Straight}       & \best{0.9564} \\
& Up-Left               & 0.9502 \\
& Three-quarter Left       & 0.9493 \\
& Three-quarter Right      & 0.9399 \\
& Down                  & 0.9363 \\
& Up                    & 0.9334 \\
& Down-Left             & 0.9301 \\
& Down-Right            & 0.9206 \\
& Up-Right              & 0.9196 \\
& Profile Left       & 0.9043 \\
& Profile Right      & 0.9041 \\
\midrule
\multirow{6}{*}{Illumination}
& \best{Ceiling + Front + Right lamps} & \best{0.9407} \\
& Front                   & 0.9403 \\
& Right diffuse lamp                   & 0.9375 \\
& Left                    & 0.9325 \\
& Outdoor                       & 0.8537 \\
& Far-right diffuse overhead lamp              & 0.8303 \\
\bottomrule
\end{tabular}
\end{table}

%% file: assets/tables/perceptual_similarity.tex
\begin{table}[t]
  \centering
  \caption{\textbf{Perceptual Similarity Comparison.} LFD reconstructions achieve higher perceptual similarity to standard webcam images than FCFD across all three metrics, indicating a smaller domain gap with respect to standard lens-based images.}
  \label{tab:perceptualsim}
  \renewcommand{\arraystretch}{1.35}
  \setlength{\tabcolsep}{5pt}
  \begin{tabular*}{\columnwidth}{@{\extracolsep{\fill}}lccc@{}}
    \toprule
    \textbf{Data/Method} & \textbf{CX $\uparrow$} & \textbf{LPIPS $\downarrow$} & \textbf{DreamSim $\downarrow$} \\
    \midrule
    FCFD (Closed-form) & 0.0392 & 0.7704 & 0.4815 \\
    FCFD (FlatNet)     & 0.1224 & 0.5451 & 0.3687 \\
    LFD (FlatNet)      & \textbf{0.2073} & \textbf{0.4829} & \textbf{0.2870} \\
    \bottomrule
  \end{tabular*}
\end{table}


%% file: assets/tables/image_quality_tpr_verification.tex
\begin{table}[t]
  \centering
  \caption{\textbf{Cross-modal TPR (\%) at fixed FPRs (\%) on the Easy subsets.} Each lensless reconstruction is paired with a standard webcam image. The LFD finetuned model yields the highest TPR across all FPRs.}
  \label{tab:tpr_verification}
  \renewcommand{\arraystretch}{1.3}
  \setlength{\tabcolsep}{4pt}
  \begin{tabular*}{\columnwidth}{@{\extracolsep{\fill}}lcccc@{}}
    \toprule
    \textbf{Model} & \textbf{FPR=0.05} & \textbf{FPR=0.1} & \textbf{FPR=0.2} & \textbf{FPR=0.5} \\
    \midrule
    Standard (baseline)   & 93.83 & 94.25 & 95.65 & 96.93 \\
    FCFD (Closed-form)    & 56.30 & 67.86 & 71.64 & 82.14 \\
    FCFD (FlatNet)        & 66.02 & 71.34 & 77.58 & 86.50 \\
    LFD (FlatNet)         & \textbf{77.30} & \textbf{80.34} & \textbf{82.78} & \textbf{86.54} \\
    \bottomrule
  \end{tabular*}
\end{table}


%% file: assets/tables/prototype2_auc.tex
\begin{table}[t]
  \centering
  \caption{\textbf{AUC Scores for PhlatCam Prototype 2 Evaluation.} The model trained on the LFD Prototype 1 training set outperforms all others across all evaluation subsets.}
  \label{tab:proto2_auc}
  \renewcommand{\arraystretch}{1.35}
  \setlength{\tabcolsep}{5pt}
  \begin{tabular*}{\columnwidth}{@{\extracolsep{\fill}}lccccc@{}}
    \toprule
    \textbf{Model} & \makecell{\textbf{Indoor}\\\textbf{Easy}} & \makecell{\textbf{Indoor}\\\textbf{All}} & \textbf{Outdoor} & \textbf{Complete} & \textbf{Hard} \\
    \midrule
    Standard  & 0.910 & 0.856 & 0.675 & 0.803 & 0.660 \\
    Simulated   & 0.923 & 0.867 & 0.678 & 0.811 & 0.677 \\
    LFD (Ours) & \textbf{0.949} & \textbf{0.912} & \textbf{0.799} & \textbf{0.882} & \textbf{0.732} \\
    \bottomrule
  \end{tabular*}
\end{table}

%% file: assets/tables/randombinary_auc.tex
\begin{table}[t]
  \centering
  \caption{\textbf{AUC Score for Random Binary Camera Evaluation.} The model trained on the LFD Prototype 1 training set outperforms all others for the Indoor Easy subset.}
  \label{tab:randombinary_auc}
  \renewcommand{\arraystretch}{1.35}
  \setlength{\tabcolsep}{6pt}
  \begin{tabular*}{0.7\columnwidth}{@{\extracolsep{\fill}}lc@{}}
    \toprule
    \textbf{Model} & \textbf{Indoor Easy} \\
    \midrule
    Standard                  & 0.8110 \\
    Random Binary Simulated   & 0.8004 \\
    FCFD-Closed               & 0.8208 \\
    FCFD-FlatNet              & 0.8409 \\
    LFD (Ours)                & \textbf{0.8563} \\
    \bottomrule
  \end{tabular*}
\end{table}



%% file: references.bib
@inproceedings{deng2017marginal,
  title={Marginal loss for deep face recognition},
  author={Deng, Jiankang and Zhou, Yuxiang and Zafeiriou, Stefanos},
  booktitle={Proceedings of the IEEE conference on computer vision and pattern recognition workshops},
  pages={60--68},
  year={2017}
}

@inproceedings{taigman2014deepface,
  title={Deepface: Closing the gap to human-level performance in face verification},
  author={Taigman, Yaniv and Yang, Ming and Ranzato, Marc'Aurelio and Wolf, Lior},
  booktitle={Proceedings of the IEEE conference on computer vision and pattern recognition},
  pages={1701--1708},
  year={2014}
}

@inproceedings{deng2019arcface,
  title={Arcface: Additive angular margin loss for deep face recognition},
  author={Deng, Jiankang and Guo, Jia and Xue, Niannan and Zafeiriou, Stefanos},
  booktitle={Proceedings of the IEEE/CVF conference on computer vision and pattern recognition},
  pages={4690--4699},
  year={2019}
}

@inproceedings{yang2020fan,
  title={Fan-face: a simple orthogonal improvement to deep face recognition},
  author={Yang, Jing and Bulat, Adrian and Tzimiropoulos, Georgios},
  booktitle={Proceedings of the AAAI Conference on Artificial Intelligence},
  volume={34},
  number={07},
  pages={12621--12628},
  year={2020}
}

@article{ren2015faster,
  title={Faster r-cnn: Towards real-time object detection with region proposal networks},
  author={Ren, Shaoqing and He, Kaiming and Girshick, Ross and Sun, Jian},
  journal={Advances in neural information processing systems},
  volume={28},
  year={2015}
}

@inproceedings{parkhi2015deep,
  title={Deep face recognition},
  author={Parkhi, Omkar and Vedaldi, Andrea and Zisserman, Andrew},
  booktitle={BMVC 2015-Proceedings of the British Machine Vision Conference 2015},
  year={2015},
  organization={British Machine Vision Association}
}

@article{asif2016flatcam,
  title={Flatcam: Thin, lensless cameras using coded aperture and computation},
  author={Asif, M Salman and Ayremlou, Ali and Sankaranarayanan, Aswin and Veeraraghavan, Ashok and Baraniuk, Richard G},
  journal={IEEE Transactions on Computational Imaging},
  volume={3},
  number={3},
  pages={384--397},
  year={2016},
  publisher={IEEE}
}

@article{tan2018face,
  title={Face detection and verification using lensless cameras},
  author={Tan, Jasper and Niu, Li and Adams, Jesse K and Boominathan, Vivek and Robinson, Jacob T and Baraniuk, Richard G and Veeraraghavan, Ashok},
  journal={IEEE Transactions on Computational Imaging},
  volume={5},
  number={2},
  pages={180--194},
  year={2018},
  publisher={IEEE}
}

@article{boominathan2020phlatcam,
  title={Phlatcam: Designed phase-mask based thin lensless camera},
  author={Boominathan, Vivek and Adams, Jesse K and Robinson, Jacob T and Veeraraghavan, Ashok},
  journal={IEEE transactions on pattern analysis and machine intelligence},
  volume={42},
  number={7},
  pages={1618--1629},
  year={2020},
  publisher={IEEE}
}

@article{antipa2017diffusercam,
  title={DiffuserCam: lensless single-exposure 3D imaging},
  author={Antipa, Nick and Kuo, Grace and Heckel, Reinhard and Mildenhall, Ben and Bostan, Emrah and Ng, Ren and Waller, Laura},
  journal={Optica},
  volume={5},
  number={1},
  pages={1--9},
  year={2017},
  publisher={Optical Society of America}
}

@article{cai2024phocolens,
  title={Phocolens: Photorealistic and consistent reconstruction in lensless imaging},
  author={Cai, Xin and You, Zhiyuan and Zhang, Hailong and Liu, Wentao and Gu, Jinwei and Xue, Tianfan},
  journal={Advances in Neural Information Processing Systems},
  volume={37},
  pages={12219--12242},
  year={2024}
}

@inproceedings{verma2025diffusion,
  title={Diffusion Model Based Image Reconstruction in Lensless Imaging},
  author={Verma, Ashish and Boominathan, Vivek and Veeraraghavan, Ashok and Seelamantula, Chandra Sekhar},
  booktitle={ICASSP 2025-2025 IEEE International Conference on Acoustics, Speech and Signal Processing (ICASSP)},
  pages={1--5},
  year={2025},
  organization={IEEE}
}

@article{khan2020flatnet,
  title={Flatnet: Towards photorealistic scene reconstruction from lensless measurements},
  author={Khan, Salman Siddique and Sundar, Varun and Boominathan, Vivek and Veeraraghavan, Ashok and Mitra, Kaushik},
  journal={IEEE Transactions on Pattern Analysis and Machine Intelligence},
  volume={44},
  number={4},
  pages={1934--1948},
  year={2020},
  publisher={IEEE}
}

@inproceedings{jain2025flattrack,
  title={FlatTrack: Eye-tracking with ultra-thin lensless cameras},
  author={Jain, Purvam and Nazar, Althaf M and Khan, Salman S and Mitra, Kaushik and Chakravarthula, Praneeth},
  booktitle={Proceedings of the Winter Conference on Applications of Computer Vision},
  pages={433--441},
  year={2025}
}

@article{khan2024opencam,
  title={Opencam: Lensless optical encryption camera},
  author={Khan, Salman S and Yu, Xiang and Mitra, Kaushik and Chandraker, Manmohan and Pittaluga, Francesco},
  journal={IEEE Transactions on Computational Imaging},
  volume={10},
  pages={1306--1316},
  year={2024},
  publisher={IEEE}
}

@article{adams2022vivo,
  title={In vivo lensless microscopy via a phase mask generating diffraction patterns with high-contrast contours},
  author={Adams, Jesse K and Yan, Dong and Wu, Jimin and Boominathan, Vivek and Gao, Sibo and Rodriguez, Alex V and Kim, Soonyoung and Carns, Jennifer and Richards-Kortum, Rebecca and Kemere, Caleb and others},
  journal={Nature Biomedical Engineering},
  volume={6},
  number={5},
  pages={617--628},
  year={2022},
  publisher={Nature Publishing Group UK London}
}

@inproceedings{yang2016wider,
  title={Wider face: A face detection benchmark},
  author={Yang, Shuo and Luo, Ping and Loy, Chen-Change and Tang, Xiaoou},
  booktitle={Proceedings of the IEEE conference on computer vision and pattern recognition},
  pages={5525--5533},
  year={2016}
}

@inproceedings{lin2014microsoft,
  title={Microsoft coco: Common objects in context},
  author={Lin, Tsung-Yi and Maire, Michael and Belongie, Serge and Hays, James and Perona, Pietro and Ramanan, Deva and Doll{\'a}r, Piotr and Zitnick, C Lawrence},
  booktitle={European conference on computer vision},
  pages={740--755},
  year={2014},
  organization={Springer}
}

@inproceedings{huiskes2008mir,
  title={The mir flickr retrieval evaluation},
  author={Huiskes, Mark J and Lew, Michael S},
  booktitle={Proceedings of the 1st ACM international conference on Multimedia information retrieval},
  pages={39--43},
  year={2008}
}

@inproceedings{cao2018vggface2,
  title={Vggface2: A dataset for recognising faces across pose and age},
  author={Cao, Qiong and Shen, Li and Xie, Weidi and Parkhi, Omkar M and Zisserman, Andrew},
  booktitle={2018 13th IEEE international conference on automatic face \& gesture recognition (FG 2018)},
  pages={67--74},
  year={2018},
  organization={IEEE}
}

@inproceedings{huang2008labeled,
  title={Labeled faces in the wild: A database forstudying face recognition in unconstrained environments},
  author={Huang, Gary B and Mattar, Marwan and Berg, Tamara and Learned-Miller, Eric},
  booktitle={Workshop on faces in'Real-Life'Images: detection, alignment, and recognition},
  year={2008}
}

@article{zheng2018cross,
  title={Cross-pose lfw: A database for studying cross-pose face recognition in unconstrained environments},
  author={Zheng, Tianyue and Deng, Weihong},
  journal={Beijing University of Posts and Telecommunications, Tech. Rep},
  volume={5},
  number={7},
  pages={5},
  year={2018}
}

@inproceedings{koestinger2011annotated,
  title={Annotated facial landmarks in the wild: A large-scale, real-world database for facial landmark localization},
  author={Koestinger, Martin and Wohlhart, Paul and Roth, Peter M and Bischof, Horst},
  booktitle={2011 IEEE international conference on computer vision workshops (ICCV workshops)},
  pages={2144--2151},
  year={2011},
  organization={IEEE}
}

@misc{nafie_headpose_github,
  author       = {Mostafa Nafie},
  title        = {Head-Pose-Estimation},
  url = {https://github.com/Mostafa-Nafie/Head-Pose-Estimation},
  year         = {2022},
  note         = {Github repository}
}

@article{lugaresi2019mediapipe,
  title={Mediapipe: A framework for building perception pipelines},
  author={Lugaresi, Camillo and Tang, Jiuqiang and Nash, Hadon and McClanahan, Chris and Uboweja, Esha and Hays, Michael and Zhang, Fan and Chang, Chuo-Ling and Yong, Ming Guang and Lee, Juhyun and others},
  journal={arXiv preprint arXiv:1906.08172},
  year={2019}
}

@inproceedings{ronneberger2015u,
  title={U-net: Convolutional networks for biomedical image segmentation},
  author={Ronneberger, Olaf and Fischer, Philipp and Brox, Thomas},
  booktitle={International Conference on Medical image computing and computer-assisted intervention},
  pages={234--241},
  year={2015},
  organization={Springer}
}

@inproceedings{hu2018squeeze,
  title={Squeeze-and-excitation networks},
  author={Hu, Jie and Shen, Li and Sun, Gang},
  booktitle={Proceedings of the IEEE conference on computer vision and pattern recognition},
  pages={7132--7141},
  year={2018}
}

@article{zhao2003face,
  title={Face recognition: A literature survey},
  author={Zhao, Wenyi and Chellappa, Rama and Phillips, P Jonathon and Rosenfeld, Azriel},
  journal={ACM computing surveys (CSUR)},
  volume={35},
  number={4},
  pages={399--458},
  year={2003},
  publisher={ACM New York, NY, USA}
}

@article{fawcett2006introduction,
  title={An introduction to ROC analysis},
  author={Fawcett, Tom},
  journal={Pattern recognition letters},
  volume={27},
  number={8},
  pages={861--874},
  year={2006},
  publisher={Elsevier}
}

@inproceedings{zhang2018unreasonable,
  title={The unreasonable effectiveness of deep features as a perceptual metric},
  author={Zhang, Richard and Isola, Phillip and Efros, Alexei A and Shechtman, Eli and Wang, Oliver},
  booktitle={Proceedings of the IEEE conference on computer vision and pattern recognition},
  pages={586--595},
  year={2018}
}

@inproceedings{mechrez2018contextual,
  title={The contextual loss for image transformation with non-aligned data},
  author={Mechrez, Roey and Talmi, Itamar and Zelnik-Manor, Lihi},
  booktitle={Proceedings of the European conference on computer vision (ECCV)},
  pages={768--783},
  year={2018}
}

@article{fu2023dreamsim,
  title={Dreamsim: Learning new dimensions of human visual similarity using synthetic data},
  author={Fu, Stephanie and Tamir, Netanel and Sundaram, Shobhita and Chai, Lucy and Zhang, Richard and Dekel, Tali and Isola, Phillip},
  journal={arXiv preprint arXiv:2306.09344},
  year={2023}
}

@article{monakhova2019learned,
  title={Learned reconstructions for practical mask-based lensless imaging},
  author={Monakhova, Kristina and Yurtsever, Joshua and Kuo, Grace and Antipa, Nick and Yanny, Kyrollos and Waller, Laura},
  journal={Optics express},
  volume={27},
  number={20},
  pages={28075--28090},
  year={2019},
  publisher={Optical Society of America}
}

@inproceedings{reshetouski2020lensless,
  title={Lensless imaging with focusing sparse ura masks in long-wave infrared and its application for human detection},
  author={Reshetouski, Ilya and Oyaizu, Hideki and Nakamura, Kenichiro and Satoh, Ryuta and Ushiki, Suguru and Tadano, Ryuichi and Ito, Atsushi and Murayama, Jun},
  booktitle={European Conference on Computer Vision},
  pages={237--253},
  year={2020},
  organization={Springer}
}

@article{fu2022diffractive,
  title={Diffractive lensless imaging with optimized Voronoi-Fresnel phase},
  author={Fu, Qiang and Yan, Dong-Ming and Heidrich, Wolfgang},
  journal={Optics Express},
  volume={30},
  number={25},
  pages={45807--45823},
  year={2022},
  publisher={Optica Publishing Group}
}

@misc{monk2023monk,
  author       = {Ellis Monk},
  title        = {The Monk Skin Tone Scale},
  year         = {2023},
  note         = {Open Science Framework},
}

@inproceedings{henry2023privacy,
  title={Privacy preserving face recognition with lensless camera},
  author={Henry, Chris and Asif, M Salman and Li, Zhu},
  booktitle={ICASSP 2023-2023 IEEE International Conference on Acoustics, Speech and Signal Processing (ICASSP)},
  pages={1--5},
  year={2023},
  organization={IEEE}
}

@article{cai2024lenslessface,
  title={LenslessFace: an end-to-end optimized lensless system for privacy-preserving face verification},
  author={Cai, Xin and Zhang, Hailong and Wang, Chenchen and Liu, Wentao and Gu, Jinwei and Xue, Tianfan},
  journal={arXiv preprint arXiv:2406.04129},
  year={2024}
}

@inproceedings{guo2016ms,
  title={Ms-celeb-1m: A dataset and benchmark for large-scale face recognition},
  author={Guo, Yandong and Zhang, Lei and Hu, Yuxiao and He, Xiaodong and Gao, Jianfeng},
  booktitle={European conference on computer vision},
  pages={87--102},
  year={2016},
  organization={Springer}
}
